\documentclass{article}
\usepackage{amsmath,amsfonts}
\usepackage{algorithmic}
\usepackage{array}
\usepackage{tikz}
\usetikzlibrary{positioning}
\usetikzlibrary{calc} 

\usepackage[margin=1in]{geometry}

\usepackage[caption=false,font=normalsize,labelfont=sf,textfont=sf]{subfig}
\usepackage{textcomp}
\usepackage{stfloats}
\usepackage{url}
\usepackage{verbatim}
\usepackage{graphicx}

\usepackage{balance}

\begin{document}
\title{Quantifying and Visualizing Sim-to-Real Gaps: Physics-Guided Regularization for Reproducibility}
\author{Yuta~Kawachi\thanks{DENSO IT Laboratory}}

\maketitle

\begin{abstract}
Simulation-to-real transfer using domain randomization for robot control often relies on low-gear-ratio, backdrivable actuators, but these approaches break down when the sim-to-real gap widens. Inspired by the traditional PID controller, we reinterpret its gains as surrogates for complex, unmodeled plant dynamics. We then introduce a physics-guided gain regularization scheme that measures a robot's effective proportional gains \((k_x, k_v, k_\theta, k_\omega)\) via simple real-world experiments. Then, we penalize any deviation of a neural controller's local input--output sensitivities from these values during training. To avoid the overly conservative bias of na\"{i}ve domain randomization, we also condition the controller on the current plant parameters. 
On an off-the-shelf two-wheeled balancing robot with a 110:1 gearbox, our gain-regularized, parameter-conditioned RNN achieves angular settling times in hardware that closely match simulation. At the same time, a purely domain-randomized policy exhibits persistent oscillations and a substantial sim-to-real gap. These results demonstrate a lightweight, reproducible framework for closing sim-to-real gaps on affordable robotic hardware.
\end{abstract}

\section{Introduction}
Simulation-to-real (sim-to-real) transfer has achieved significant success on robotic platforms equipped with low-gear-ratio, backdrivable actuators. In these systems, such as small humanoids that learn to walk, reinforcement learning (RL) combined with domain randomization (DR) often closes the gap between virtual training and physical deployment \cite{rudin2022learning}\cite{cha2025sim}\cite{zhang2025hub}. By exposing the controller to modest parameter perturbations in simulation, these methods can operate on real hardware without detailed modeling of friction, backlash, or sensor delay \cite{katayama2025learning}.

However, this success does not readily extend to tasks with a wider sim-to-real gap \cite{mahajan2024quantifying} \cite{Waheed10904321}. High-gear-ratio mechanisms \cite{katayama2025learning}, tight object-insertion assemblies \cite{marougkas2025integrating}, and aggressive maneuvers such as parkour jumps \cite{chane2024soloparkour} or skateboarding \cite{liu2025discrete} require additional hardware or training tricks. When the randomization range in simulation is enlarged by these requirements,
\begin{enumerate}
    \item the probability that a sampled model matches the real plant decreases, which lowers the \emph{hit rate}, and
    \item the learned policy is marginalized over many inaccurate models, which yields the ``overly conservative policies'' problem \cite{he2025asap} \cite{he2024learning}.
\end{enumerate}
From a scientific standpoint, the RL+DR methodology is closer to gambling than engineering: an RL agent \emph{gets lucky} if it converges to a policy that transfers. In industrial settings, in which reproducibility and certifiable performance are critical, depending on serendipity is unacceptable.
Therefore, closing significant sim-to-real gaps requires more accurate knowledge of the physical system.

A straightforward remedy is \emph{real-to-sim} system identification \cite{zheng2023close} or iterative simulator refinement. Although this approach can improve fidelity, it demands extensive data collection limited by safety considerations and hardware wear. Furthermore, any unmeasured or hazardous operating regime remains poorly modeled, so the refined simulator can still diverge from reality when the robot executes aggressive motions.

Another option is to abandon physics modeling altogether and \emph{gauge} the hardware directly. The only practical controller family that is both human-interpretable and suitable for direct gauging is the classical proportional-integral-derivative (PID) controller. Nevertheless, PID control is inherently linear and cannot deliver the nonlinear performance required for wide-amplitude swing-up, contact-rich manipulation, or motion with large gear reductions.

In this work, we adopt a different perspective. We reinterpret proportional gains, which are traditionally treated as human-tuned design knobs, as surrogates for the complex nature of unmodeled plant characteristics, encapsulating bandwidth, friction, and gear-train dynamics. We incorporate \emph{physics-guided gain regularization} into the learning loop. We measure the robot's effective proportional gains
\(
(k_{x},\,k_{v},\,k_{\theta},\,k_{\omega})
\)
and penalize any deviation of a neural controller's local input--output sensitivities from these hardware gains during training. This mechanism prevents the simulator from rewarding unrealistically large virtual gains and keeps the learned policy within the feasible actuation envelope of the real machine while maintaining the theoretical nonlinear control capability.

To counteract the ``overly conservative policies'' that emerge from na\"{i}ve DR, we introduce \emph{parameter-conditioned domain randomization}. The current plant parameters are appended to the policy input so the network can specialize its output instead of collapsing toward a single compromise solution. The combination of physics-guided gain regularization and parameter conditioning delivers reproducible, high-performance control on an off-the-shelf two-wheeled balancing robot with a 110:1 gearbox, which demonstrates that lightweight physics guidance, rather than enlarging randomization ranges, is the practical key to affordable robotics.

In summary, our contributions are as follows.
\begin{enumerate}
  \item \textbf{Physics-guided gain regularization}: we formulate a loss that forces the input--output partial derivatives of the neural controller to match hardware gains, which prevents the simulator from rewarding unrealistically large virtual gains.
  \item \textbf{Parameter-conditioned domain randomization}: by feeding the plant parameters into the policy, we remove the bias toward overly conservative behaviors. 
  \item \textbf{Quantitative analysis of the sim-to-real gap}: we visualize how unconstrained policies inflate equivalent gains and we compare both simulated and real responses, which clearly shows the sim-to-real discrepancy in na\"{i}ve DR. 
  \item \textbf{Hardware validation on a low-cost robot}: We verified our method on a low-cost, off-the-shelf, two-wheeled, high-gear-ratio (110:1) balancing robot Balboa 32U4, which is considered to be a hard example for the na\"{i}ve RL+DR approaches. 
\end{enumerate}

\section{Method}
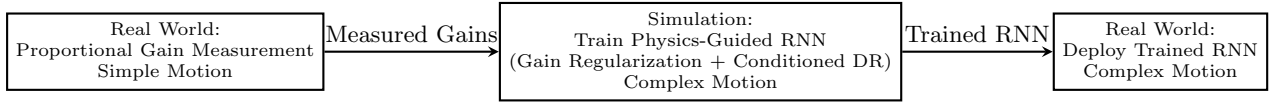
\begin{figure}[t]
  \centering
\begin{tikzpicture}[
      block/.style = {
        draw,
        rectangle,
        thick,
        fill=white,
        minimum width=2cm,
        minimum height=0.8cm,
        align=center,
        font=\scriptsize
      },
      arrow/.style={thick,->,>=stealth},
      node distance=3cm and 2cm
    ]

    \node[block] (preexp) 
      {Real World:\\
       Proportional Gain Measurement\\
       Simple Motion};

    \node[block] (simdr) at ($(preexp.east)+(5cm,0)$) 
      {Simulation:\\
       Train Physics-Guided RNN\\
       \scriptsize (Gain Regularization + Conditioned DR)\\
       \scriptsize Complex Motion};

    \node[block, right=of simdr] (deploy) 
      {Real World:\\
       Deploy Trained RNN\\
       Complex Motion};

    \draw[arrow] (preexp.east) -- node[above,font=\small] {Measured Gains} (simdr.west);
    \draw[arrow] (simdr.east) -- node[above,font=\small] {Trained RNN} (deploy.west);

  \end{tikzpicture}
  \caption{Training Pipeline for RNN Controller with Physics-Guided Regularization}
  \label{fig:training_pipeline}
\end{figure}

\subsection{Modeling}
We consider a two-wheeled balancing robot as a surrogate for a low-cost robotics platform. We adopt a planar cart-pole model with no wheel slip. This simplification captures key nonlinear control challenges while enabling analytical modeling.

The nonlinear state-space model is defined as follows \cite{CDGreen}:
\begin{align}
\Delta &= M(ml^2+J)-(ml\cos\theta)^2, \\
\dot{x} &= v, \\
\Delta \dot{v} &= -(ml)^2g\sin\theta\cos\theta + \nonumber \\
&(ml^2+J)(f + ml\omega^2\sin\theta - D_c v) + ml D_p \omega \cos\theta, \\
\dot{\theta} &= \omega, \\
\Delta \dot{\omega} &= Mmgl\sin\theta - \nonumber \\
&ml\cos\theta(f + ml\omega^2\sin\theta - D_c v) - M D_p \omega,
\end{align}

where $x$, $v$, $\theta$, $\omega$ denote horizontal position, velocity, angle, and angular velocity. $m$, $m_c$ are the pole and cart masses, $M = m + m_c$, $l$ is pole length, $J$ is its inertia, $g$ is gravity, and $D_c$, $D_p$ are viscous damping terms. We treat all friction as viscous as an engineering approximation. We use nominal parameters based on empirical tuning (Table~\ref{tab:common_parameters}).
\subsection{Controller Design and Training}

We adopt a single-layer Elman RNN controller based on \cite{kawachi2024} to handle the plant's nonlinear dynamics with minimal computation. It receives the plant state $\mathbf{x} = [x, v, \theta, \omega]^\top$ and hidden state $\mathbf{h}$, and outputs force $f$:
\begin{align}
    f = G(\mathbf{x}, \mathbf{h}) = c_{\rm out} \cdot C(\sigma(A\mathbf{h} + B\mathbf{x} + b_1)) + b_2,
\end{align}
with trainable weights $A, B, C, b_1, b_2$, and activation $\sigma$. The RNN hidden state $\mathbf{h}$ is initialized to zeros. \(c_{\rm out}\) scales the NN output to match the force range. The matrix $A$ is decomposed as in \cite{kawachi2024} to suppress the divergence of the training loss.
Unlike deep RL, we exploit known dynamics and define direct losses for control objectives. The state tracking loss is
\begin{align}
\label{eq:loss}
{\mathcal L}_s &= \mathbb{E}_{b,t} \Bigl| ({\hat s} - s)/s_{\rm base} \Bigr|
\end{align}
where $s \in \{x, v, \theta, \omega\}$, $\hat{s}=[0,0,0,0]^\top$ (means upright and stationary), The operator $\mathbb{E}_{b,t}$ denotes the average over mini-batches and time. States are normalized by reference scales $s_{\rm base}$ in the loss calculation.

To stabilize training, we apply a continuous curriculum learning (CCL) strategy as shown in Table \ref{tab:CCL}: the initial training phase uses short horizons and low noise. The simulation length, noise levels, and parameter variations gradually increase as learning proceeds. Plant parameters are randomized in the fine-tuning phase.

At the initial time step, Gaussian noise is added to the initial state variables (e.g. $x \leftarrow x_{\rm noise (scale)} \cdot \mathcal{N}(0,1)$). The cart-pole cannot move beyond the max values (clamped so that $|x| \leq x_{max}$). When we get the output force $f$ from a controller, a time-consistent external force $f_{\rm noise(scale)} \cdot {\mathcal N}(0,1)$ is added to increase robustness. The total force is also clamped so that $|f| \leq f_{max}$. The system is discretized using RK4.

\begin{table}[ht]
\centering
\caption{Continuous Curriculum Learning Procedure}
\label{tab:CCL}
\begin{tabular}{l l l}
\hline
\textbf{Parameter} & \textbf{Pretrain} & \textbf{Finetune} \\
\hline
\multicolumn{3}{l}{\textbf{Simulation Settings}} \\
\hline
epochs       & \(0\le n<50\)  & \(50\le n<1000\) \\
learn. rate        & \(5\times10^{-4}\) & \(3\times10^{-4}\) \\
$t_{\rm sim}/t_{\rm dt}$         & \(50\) & \(\min\Bigl(\lfloor (n-50)\cdot\frac{500-50}{950}+50\rfloor,\,500\Bigr)\) \\
$\theta_{\rm noise(scale)}$  & \(0.1\) & \(\min\Bigl((n-50)\cdot\frac{0.6-0.1}{950}+0.1,\;0.6\Bigr)\) \\
$f_{\rm noise(scale)}$    & 0 & 1 \\
\hline
\multicolumn{3}{l}{\textbf{Plant Parameter Domain Randomization}} \\
\hline
\(M\)               & $M_{\rm nom}$ & \(\mathcal{U}(0.5,2.0)\cdot\mathrm{M_{\rm nom}}\) \\
\(m\)               & $m_{\rm nom}$ & \(0.8\,\mathcal{U}(0.5,1.0)\cdot M\) \\
\(l\)               & $l_{\rm nom}$ & \(\mathcal{U}(0.5,2.0)\cdot\mathrm{l_{\rm nom}}\) \\
\(D_c\)              & $\mathrm{D_{c\,{\rm nom}}}$ & \(\mathcal{U}(0.5,2.0)\cdot\mathrm{D_{c\,{\rm nom}}}\) \\
\(D_p\)              & $\mathrm{D_{p\,{\rm nom}}}$ & \(\mathcal{U}(0.5,2.0)\cdot\mathrm{D_{p\,{\rm nom}}}\) \\
\hline
\end{tabular}
\label{tab:training_schedule}
\end{table}

\subsection{Physics-Guided Regularization}

\paragraph*{Gain-Constrained Learning}
Classical proportional control uses:
\begin{align}
    f_{\rm prop} = -k_x x - k_v v - k_\theta \theta - k_\omega \omega,
\end{align}
where $k_s$ are empirically tunable gains. 
We originated the idea of embedding this classical structure into the learning process to guide the RNN-based controller using the same methodology developed for solving partial differential equations \cite{PINNs}, which is typically used straightforwardly as a model of motion in the context of control \cite{giacomuzzo2024black}.
Using automatic differentiation, we can compute a gain-equivalent of the RNN:
\begin{align}
    \left[\frac{\partial f}{\partial x}, \frac{\partial f}{\partial v}, \frac{\partial f}{\partial \theta}, \frac{\partial f}{\partial \omega}\right],
\end{align}
and minimize their deviation from experimentally observed gains $k_s$:
\begin{align}
    \mathcal{L}_{\text{grad}} = \frac{1}{4} \sum_s \left| \mathbb{E}_{b,t} \left[ \frac{\partial f}{\partial s} \right] - k_s \right|,
    {\mathcal L}_{\text{total}} &= \sum_s{\mathcal {\mathcal L}_s} + {\mathcal L}_{\text{grad}}.
\end{align}
This constraint implicitly reflects real-world actuation characteristics and forces the RNN to align with them.

\paragraph*{Conditioned Domain Randomization}
We apply domain randomization over plant parameters (mass, length, and damping) to generalize across hardware uncertainty. However, na\"ive domain randomization leads to ``overly conservative policies''~\cite{he2025asap}, due to averaging over all variations.
To mitigate this, we explicitly condition the controller on the current plant parameters, which are appended to the input state. This allows the network to avoid overly conservative policies.

\paragraph*{Unified View}
The gain constraint anchors the controller's sensitivity within physically meaningful bounds, while parameter conditioning enables adaptation across plant variations. Together, they suppress overgeneralization from domain randomization and preserve robustness near nominal parameters, forming a lightweight yet effective regularization scheme for sim-to-real transfer. The total training pipeline is shown in Fig. \ref{fig:training_pipeline}.

\subsection{Deployment setting}
Sensor smoothing (exponential moving average with smoothing factor 0.2) is applied to velocities.
We formulated a simplified DC motor model:
\begin{align}
    f(t) = c_{\rm emp} \cdot \frac{c_{\mathrm{gear}}}{r_{\rm wheel}}\cdot \frac{\Phi_{\rm PM}}{R_{\rm winding}} V_{\rm bat}(t) u_{\rm duty}(t),
\end{align}
where \(u_{\mathrm{duty}}(t)\) is the signed PWM duty ratio, \(V_{\mathrm{bat}}(t)\) is the battery voltage, \(\Phi_{\rm PM}\) is the magnetic flux, \(R_{\rm winding}\) is the winding resistance, \(c_{\mathrm{gear}}\) is the total gear ratio, and \(r\) is the wheel radius. Each constant is determined from datasheets or rough estimates and $c_{\rm emp}$ absorbs modeling error. The $c_{\rm emp}$ values are 1.2 (proportional control), 2.0 (only domain-randomized RNN), and 1.5 (gain-constrained domain-randomized RNN). We use the reverse function on the output of the controller RNN. 

\begin{table}[ht]
\centering
\caption{Common Parameters}
\begin{tabular}{l l l}
\hline
\textbf{Parameter}       & \textbf{Value}       & \textbf{Unit} \\
\hline
$c_{\rm out}$            & 300                  & N \\
$f_{\rm max}$            & 20                   & N \\
$x_{\rm max}$            & 10                   & m \\
$t_{\rm sim}$            & $\leq 5$             & s \\
$t_{\rm dt}$             & 10                 & ms \\
$n_{\rm batch}$          & 16                   & -- \\
$n$                      & $\leq 2001$          & -- \\
$n_{\rm hidden\_rnn}$    & 256                  & -- \\
$x_{\rm noise (scale)}$        & 0.01                 & m \\
$v_{\rm noise (scale)}$        & 0.01                 & m/s \\
$\theta_{\rm noise (scale)}$        &  $ \leq \displaystyle {35\pi}/{180}$ & rad \\
$\omega_{\rm noise (scale)}$        & 0.01                 & rad/s \\
$x_{\rm base}$                & 2                    & m \\
$v_{\rm base}$                & 5                    & m/s \\
$h_{\rm base}$           & $2\pi$               & rad \\
$w_{\rm base}$                & $2\pi$               & rad/s \\
$M_{\rm nom}$    & 0.4   & kg   \\
$m_{\rm nom}$    & 0.3   & kg   \\
$l_{\rm nom}$    & 0.05  & m    \\
$g_{\rm nom}$    & 9.8   & m/s$^2$ \\
$D_{c\,{\rm nom}}$  & 3     & kg/s\\
$D_{p\,{\rm nom}}$  & 0.007 & Nms/rad\\
$c_{\rm gear}$ & 10032/91 $\simeq$ 110.2 & -- \\
$R_{\rm winding}$ & 4.3 & $\Omega$ \\
$\Phi_{PM}$ & $\simeq$ 2.7$\times 10^{-3}$ & Wb \\
$r_{\rm wheel}$ & 38.25$\times 10^{-3}$ & m \\
$l_{\rm wheel}$ & $2\pi \times r_{\rm wheel}$ & m \\
\hline
\end{tabular}
\label{tab:common_parameters}
\end{table}

\section{Experimental Condition}

\subsection{Hardware Setup}
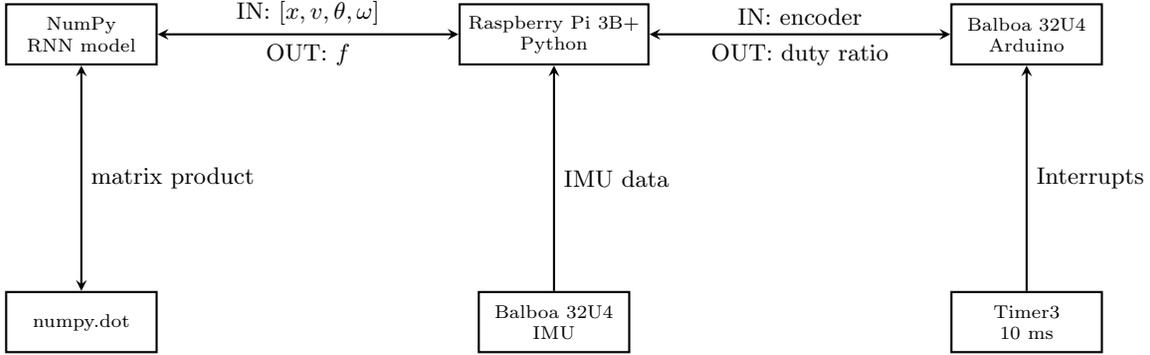
\begin{figure}[t]
  \centering
  \begin{tikzpicture}[
      block/.style = {
        draw,
        rectangle,
        thick,
        fill=white,
        minimum width=2cm,
        minimum height=0.8cm,
        align=center,
        font=\scriptsize
      },
      arrow/.style={thick,->,>=stealth},
      node distance=3cm and 2cm
    ]

    \node[block] (rnn) {NumPy\\RNN model};

    \node[block, below=of rnn] (dot) {numpy.dot};

    \node[block, right=4cm of rnn] (pi) {Raspberry Pi 3B+\\Python};
    
    \node[block, below=of pi] (imu) {Balboa 32U4\\IMU};

    \node[block, right=4cm of pi] (ard) {Balboa 32U4\\Arduino};

    \node[block, below=of ard] (timer) {Timer3\\10 ms};

    \draw[thick,<->,>=stealth]
      (dot.north) -- 
      node[right,font=\small] {matrix product} 
      (rnn.south);

    \draw[thick,<->,>=stealth]
      (rnn.east) -- 
      node[above,font=\small] {IN: $[x, v, \theta, \omega]$} 
      node[below,font=\small] {OUT: $f$} 
      (pi.west);

    \draw[arrow]
      (imu.north) -- 
      node[right,font=\small] {IMU data} 
      (pi.south);

    \draw[thick,<->,>=stealth]
      (pi.east) -- 
      node[above,font=\small] {IN: encoder} 
      node[below,font=\small] {OUT: duty ratio} 
      (ard.west);

    \draw[arrow]
      (timer.north) -- 
      node[right,font=\small] {Interrupts} 
      (ard.south);

  \end{tikzpicture}
  \caption{Control system architecture: state estimates (position, velocity, angle, angular velocity) are computed from the Balboa 32U4 IMU and encoder, sent to a NumPy RNN on a Raspberry Pi 3B+ (via numpy.dot), and the resulting duty ratio is fed back to the Balboa Arduino at 10 ms intervals using Timer3 interrupts.}
  \label{fig:arch}
\end{figure}

\begin{figure}[!t]
\centering
\includegraphics[width=2.5in]{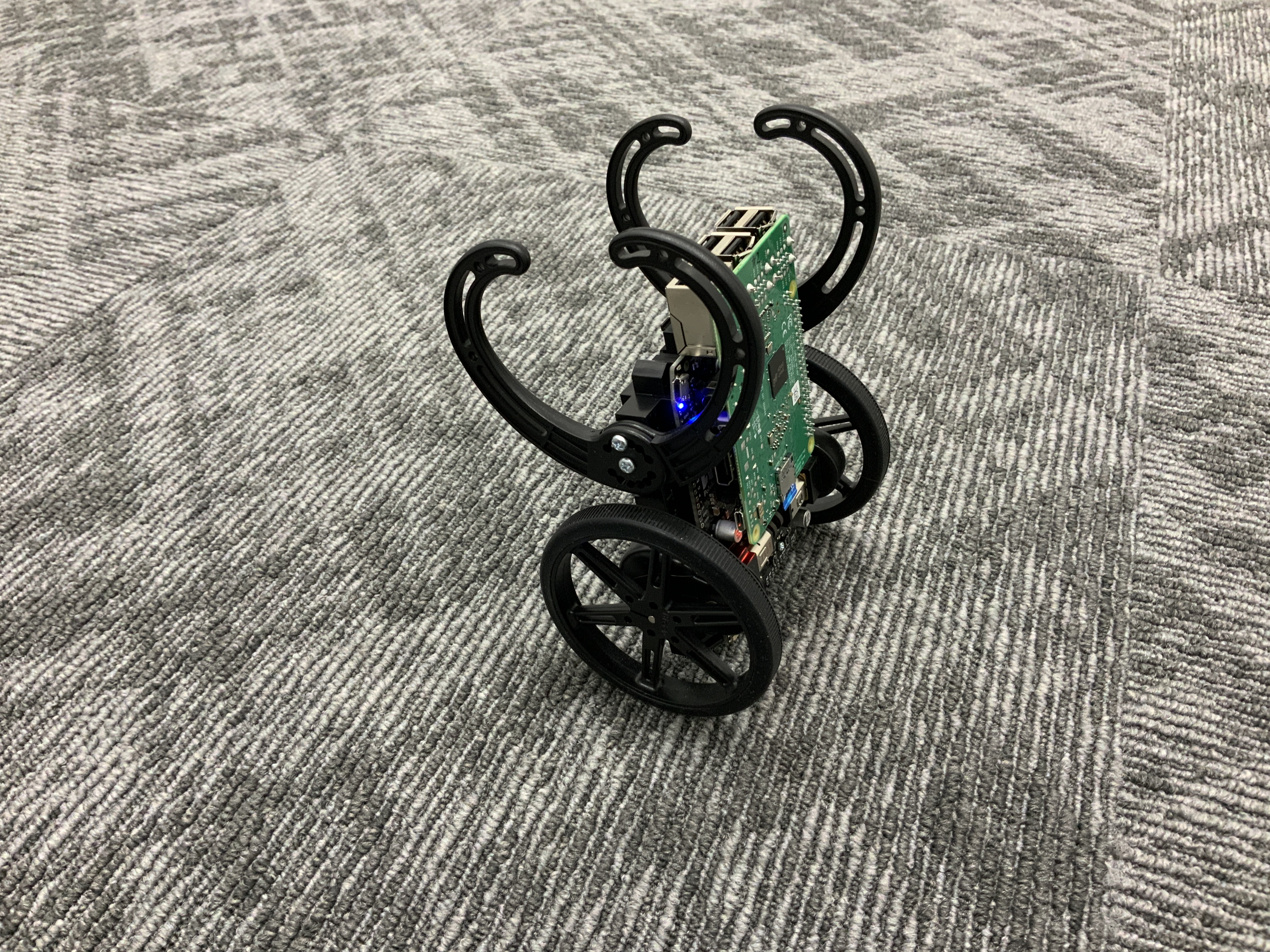}
\caption{Physical experimental platform: a two-wheeled inverted-pendulum robot (Balboa 32U4) equipped with protective arms and a Raspberry Pi 3B+ on a carpeted surface.}
\label{fig:photo}
\end{figure}

We bought the ``Balboa 32U4 Balancing Robot Kit'' with 50:1 gearboxes (item numbers 3575, 3073, and 1430) \cite{Balboa32U4}. We assembled it with Raspberry Pi 3B+. The final gear ratio is chosen as 110.2. The basic properties are shown in Table \ref {tab:common_parameters}. The system architecture is shown in Fig. \ref{fig:arch} and Fig. \ref{fig:photo}. 

\subsection{Device condition evaluation}
The reference value of the matrix-vector multiplication is 0.168 ms, and the whole routine, including the signal processing on Raspberry Pi, is 2.76 ms (without raw sensor I/O and communication bottleneck between two boards). Due to communication implementation, the actual control period varies from 10 ms to 30 ms. However, we observed that the degradation in control performance is not visible to humans, while it becomes evident when even slightly mismatched control weights are used. The battery voltage was 7.6 V to 8.0 V throughout the experiments.

\subsection{Pre-experiments}
We empirically determined the nominal proportional gains that were used in gain regularization in the loss function Equation \ref{eq:loss}. The upper and lower ranges were obtained from trial-and-error experiments on a real machine. 

\begin{table}[ht]
\centering
\caption{Working Proportional Gains from pre-experiment}
\begin{tabular}{lccccc}
\hline
\textbf{Gains} & \textbf{Lower Bound} & \textbf{Nominal} & \textbf{Upper Bound} & \textbf{Unit} \\
\hline
$k_x$           & $<$1                & 13                    & $\geq 20$  & N/m \\
$k_v$       & 12                  & 15                    & 17                   &  N/(m/s) \\
$k_\theta$          & 25                  & 31                    & 40                   &  N/rad \\
$k_\omega$     & 1.3                 & 1.6                   & 2.0--3.0             & N/(rad/s) \\
\hline
\end{tabular}
\label{tab:proportional_gains_preexperiment}
\end{table}

\section{Simulated Results}
\begin{figure}[!t]
\centering
\includegraphics[width=\linewidth]{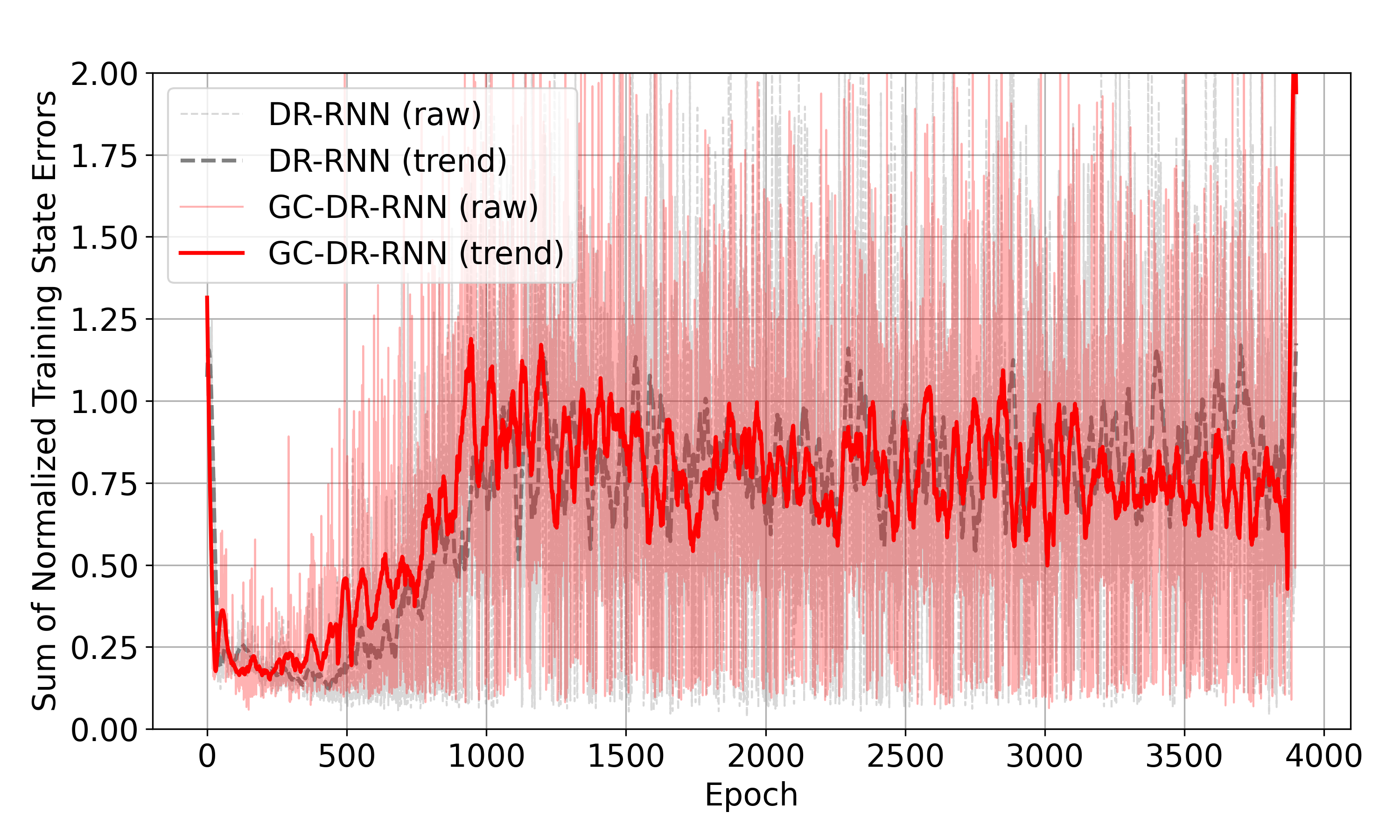}
\caption{Normalized training loss over 4000 epochs for two controllers: the domain-randomized RNN (gray) and the gain-constrained, domain-randomized RNN (red). Thin lines show raw loss; thick lines show smoothed trends. Both models converge to similar loss levels.}
\label{fig:train_loss}
\end{figure}

\begin{figure}[!t]
\centering
\includegraphics[width=\linewidth]{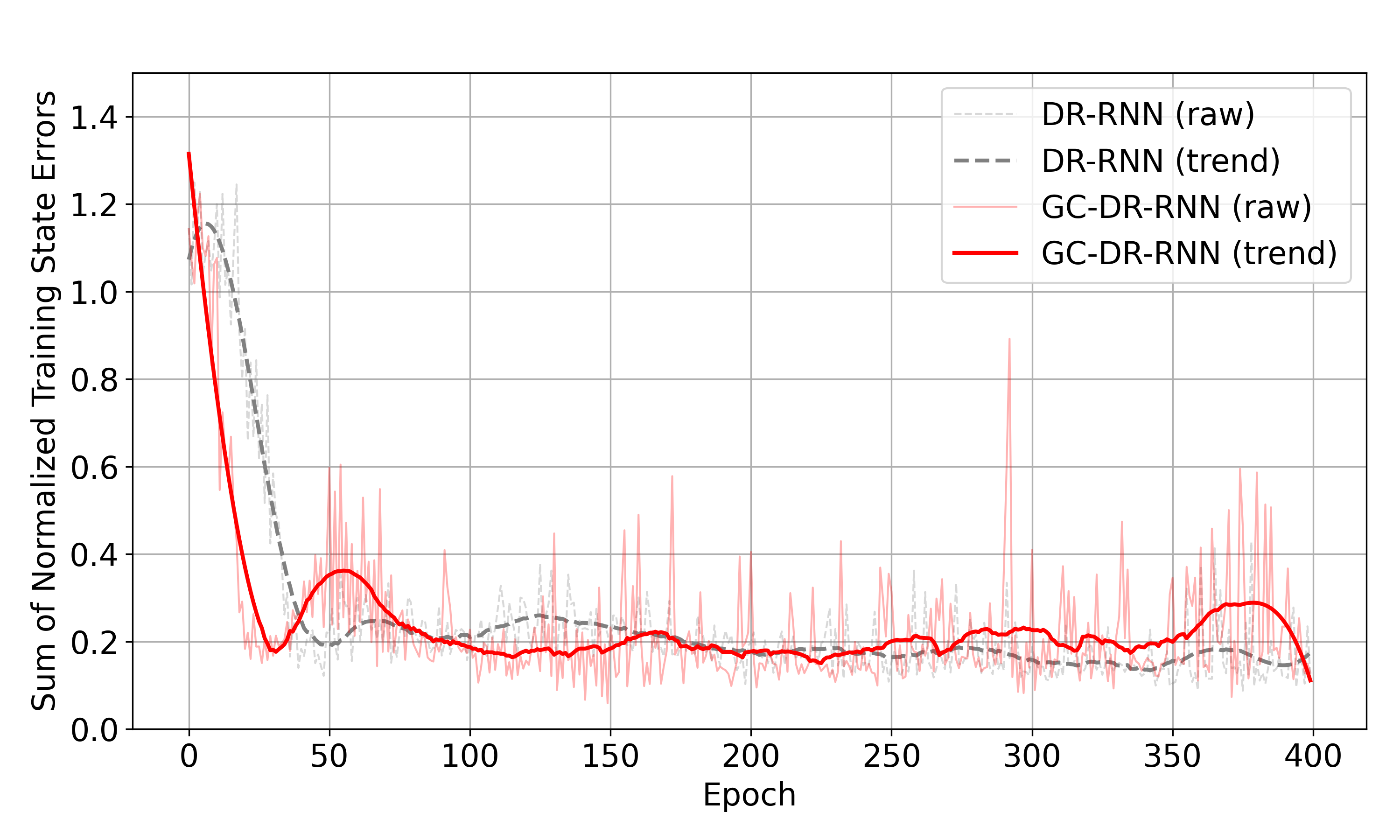}
\caption{Zoom into the first 400 epochs of the normalized training loss. The GC-DR-RNN (red) converges more quickly than the only DR-RNN (gray), demonstrating faster acquisition of basic inverted-pendulum control strategies.}
\label{fig:train_loss_400}
\end{figure}

\begin{figure}[!t]
\centering
\includegraphics[width=\linewidth]{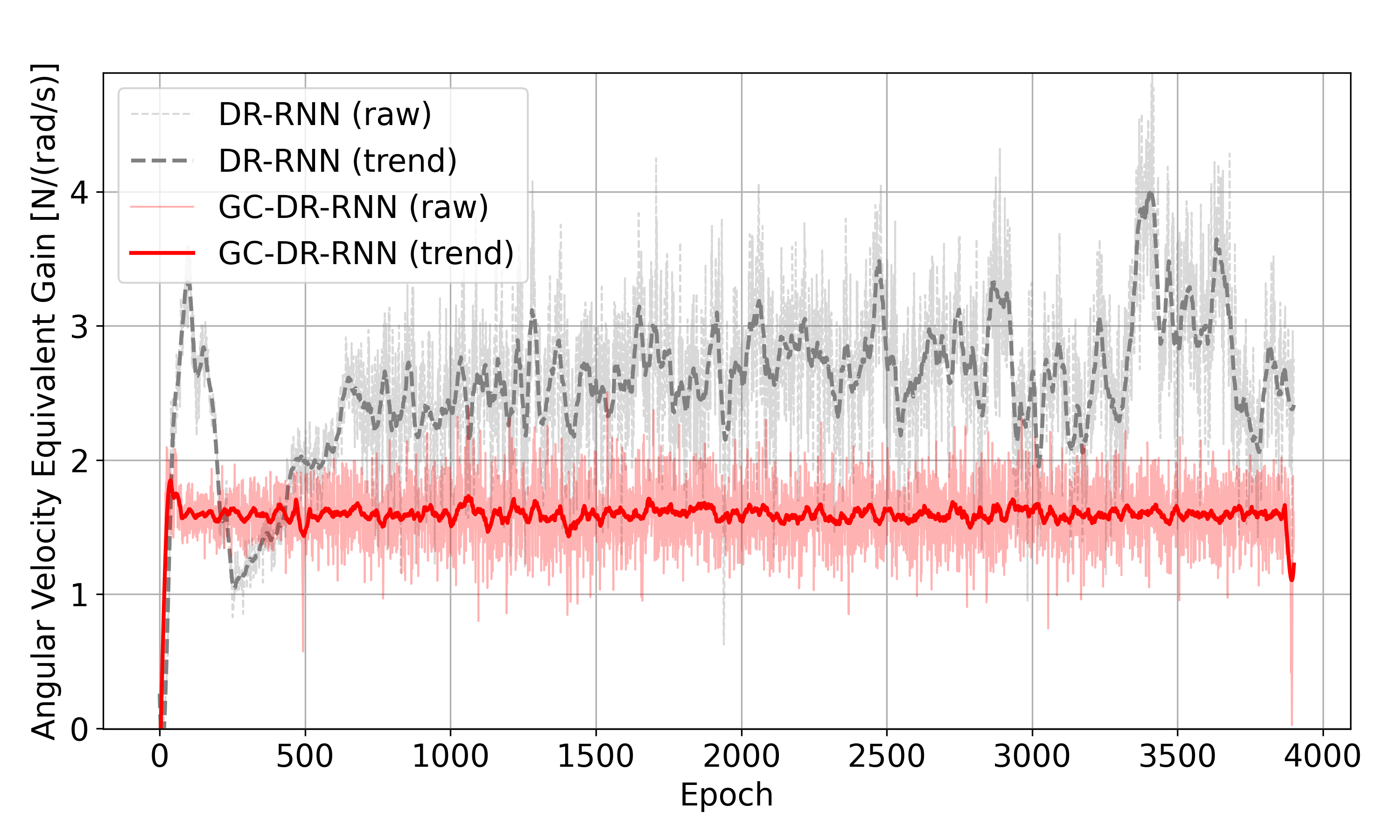}
\caption{Suppression of sim-specific equivalent-gain $\frac{\partial f}{\partial \omega}$ overfitting through regularization.}
\label{fig:gain_curves}
\end{figure}

We first compared the training loss $\sum_s{\mathcal L_s}$ curves, as shown in Fig. \ref{fig:train_loss}. We truncated both domain-randomized (DR) RNNs with or without gain constraint (GC), at 3900 training epochs. The two models achieve nearly identical training losses computed from simulated state errors.
A close-up of the curves over the initial epochs is shown in Fig. \ref{fig:train_loss_400}. The GC model converges faster than the baseline DR-RNN. We observed that this corresponds to the faster acquisition of the fundamental inverted pendulum control strategies.
The equivalent gains $\mathbb{E}_{b,t}\frac{\partial f}{\partial s}$ are shown in Fig. \ref{fig:gain_curves}. The effect of gain constraining on the GC model is evident in these results. 
In the GC model, the gains remain below 2.0, whereas those of the baseline DR-RNN fluctuate between roughly 1.0 and 4.0.

\begin{figure}[!t]
\centering
\includegraphics[width=\linewidth]{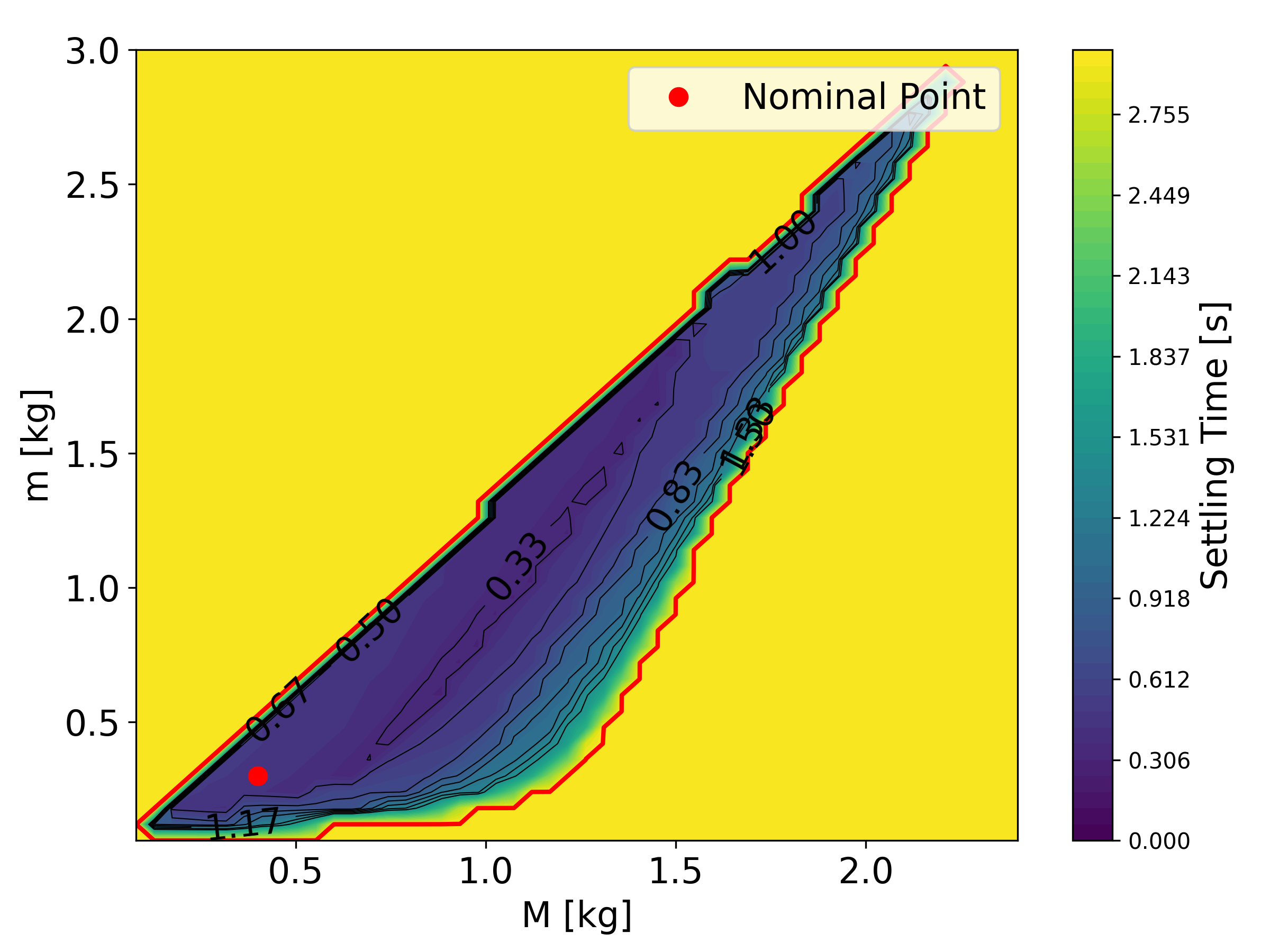}
\caption{5\% settling-time contours of the proportional controller over the cart mass \(M\) vs. pole mass \(m\) plane. Fast convergence (dark blue) is confined to a narrow band around the nominal parameters, indicating limited robustness to parameter variations.}
\label{fig:a5Mm_P}
\end{figure}

\begin{figure}[!t]
\centering
\includegraphics[width=\linewidth]{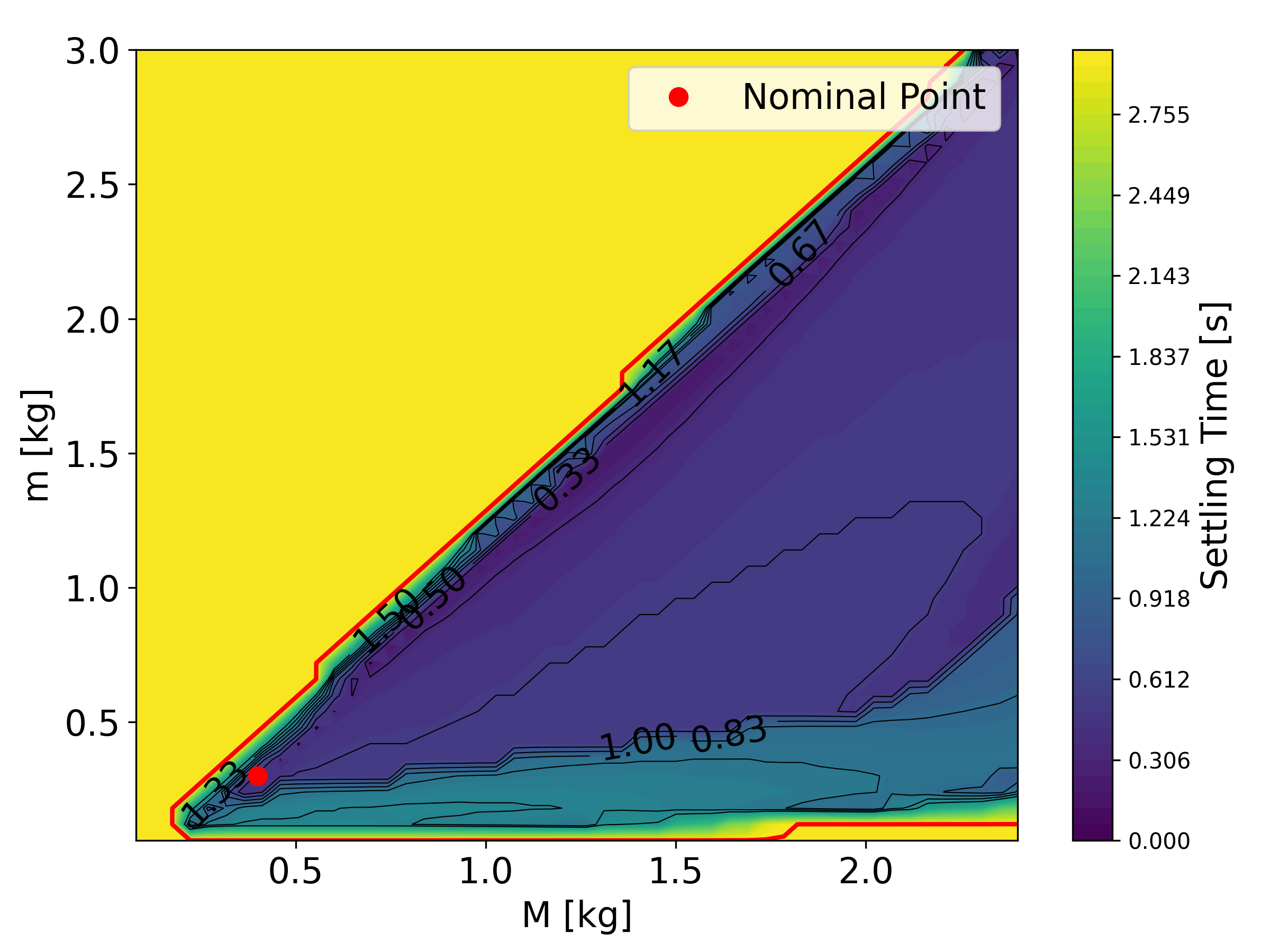}
\caption{5\% settling-time contours of the DR-RNN over the cart mass \(M\) vs. pole mass \(m\) plane: the stable region is substantially wider than the proportional baseline, but the nominal point lies near the edge of the feasible zone, suggesting reduced local stability.}
\label{fig:a5Mm_DR}
\end{figure}

\begin{figure}[!t]
\centering
\includegraphics[width=\linewidth]{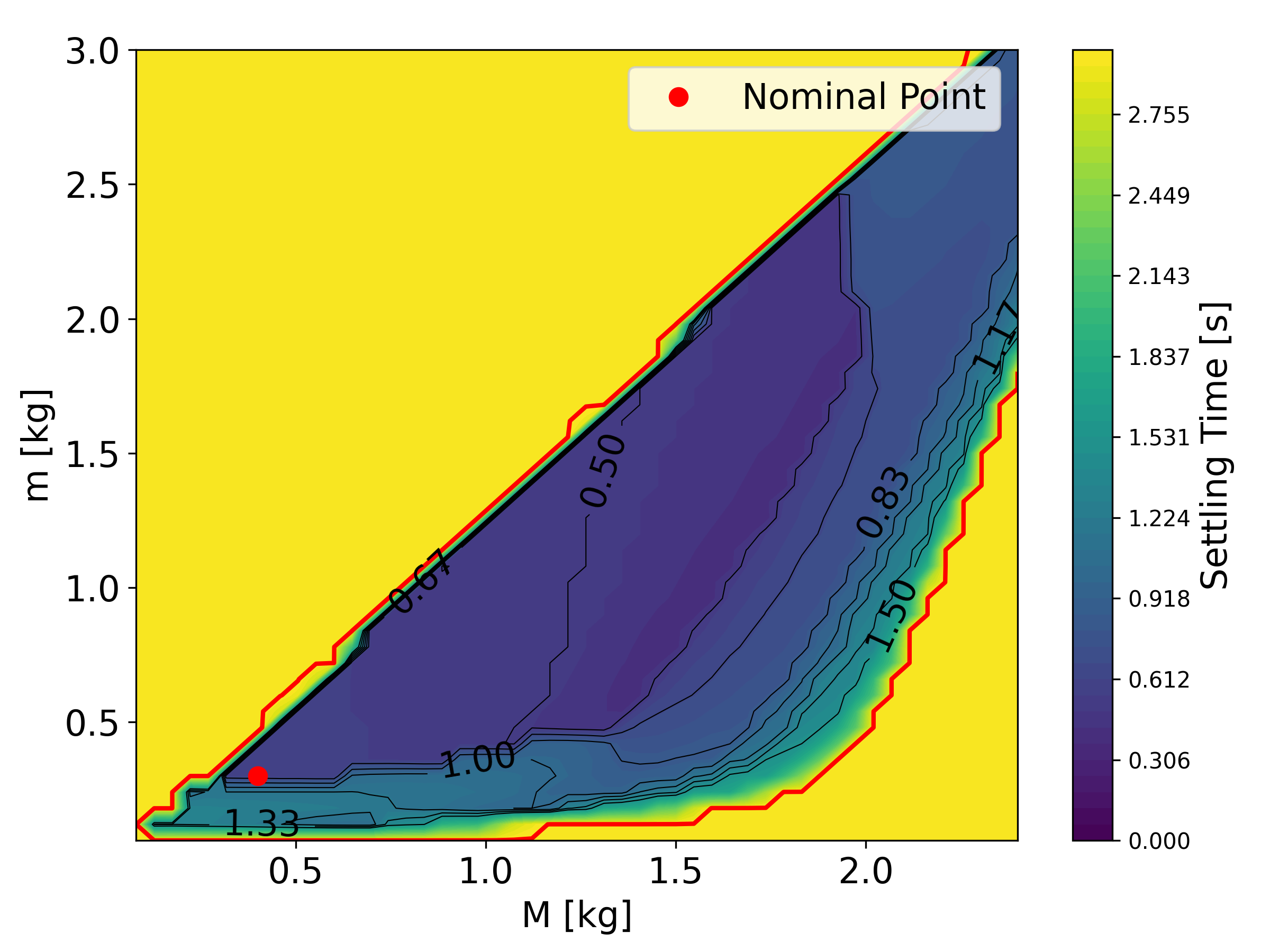}
\caption{5\% settling-time contours for the GC-DR-RNN over the cart mass \(M\) vs. pole mass \(m\) plane. A broad dark-blue region surrounds the nominal point, showing that this model balances global robustness with high performance at the true hardware parameters.}
\label{fig:a5Mm_GCDR}
\end{figure}

\begin{figure}[!t]
\centering
\includegraphics[width=\linewidth]{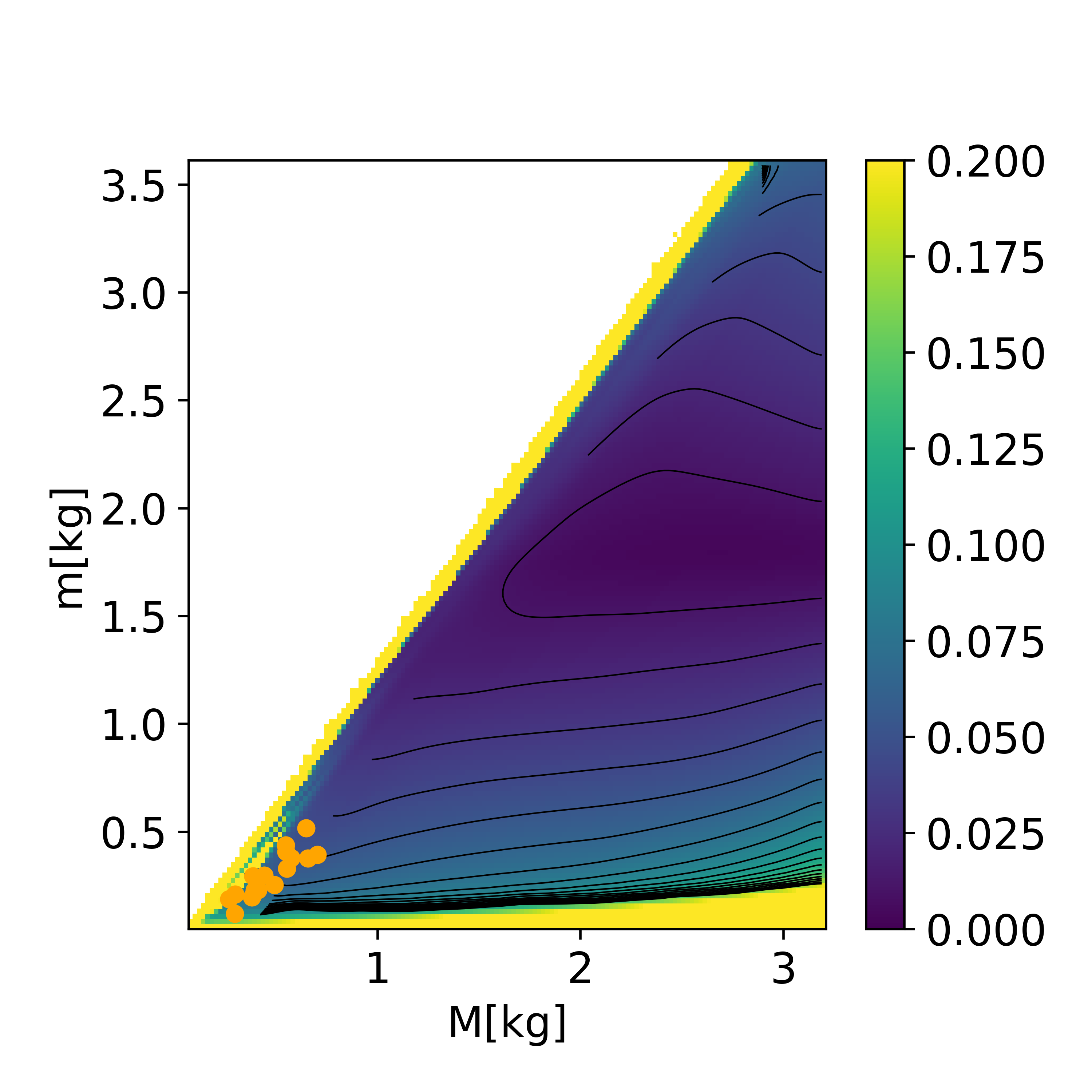}
\caption{Final minibatch sample locations (orange dots) overlaid on the contour map of angular velocity loss $\mathcal{L}_\omega$ in the $M$–$m$ plane for the DR-RNN. The model's training is biased toward “easy” low-loss regions that lie away from the true hardware parameters.}
\label{fig:wloss_DR_Mm}
\end{figure}

\begin{figure}[!t]
\centering
\includegraphics[width=\linewidth]{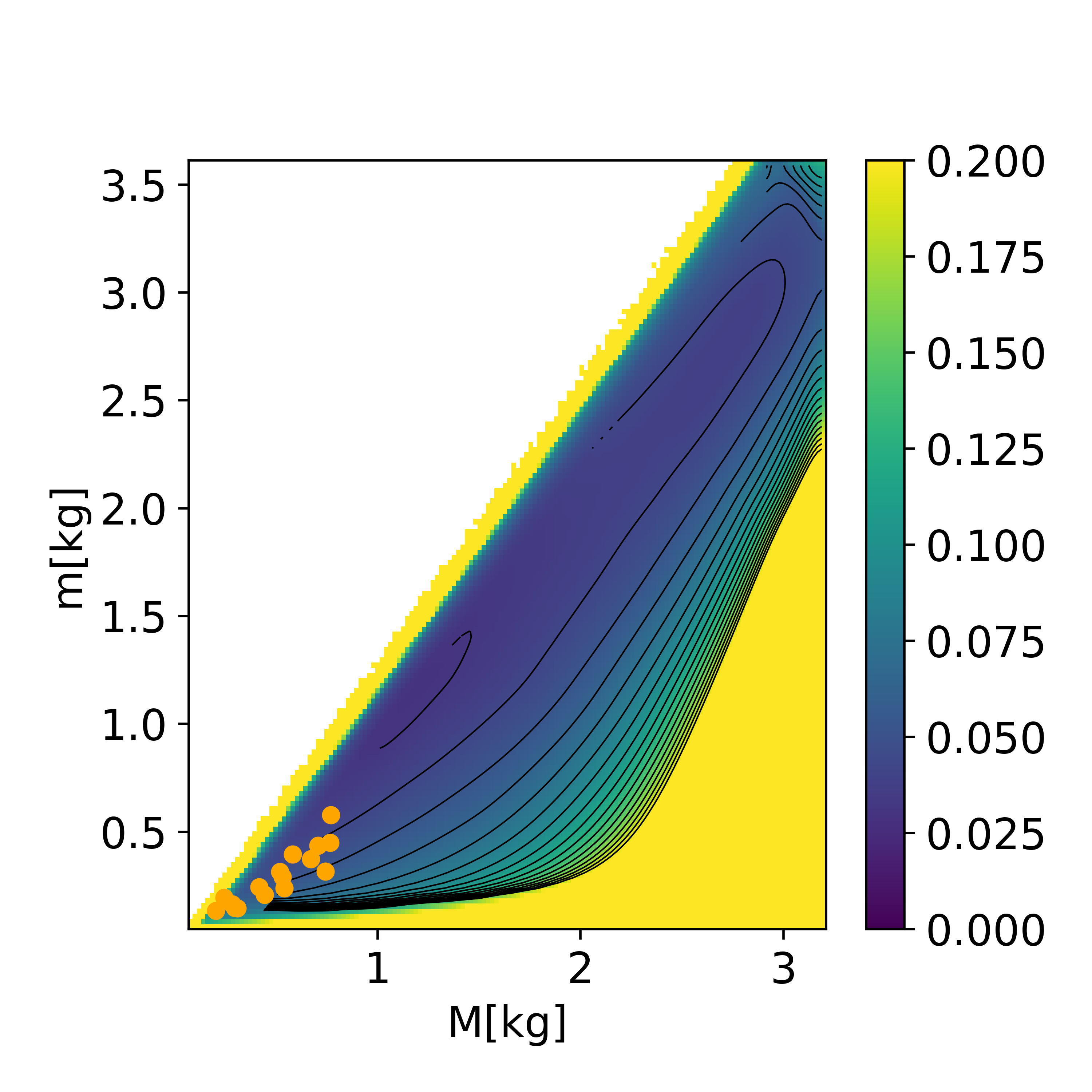}
\caption{Final minibatch sample locations (orange dots) overlaid on the contour map of angular velocity loss $\mathcal{L}_\omega$ in the $M$–$m$ plane for the GC-DR-RNN. The loss minimum is centered near the nominal hardware parameters, demonstrating suppression of drift toward unrealistic configurations.}
\label{fig:wloss_GCDR_Mm}
\end{figure}

\begin{figure}[!t]
\centering
\includegraphics[width=\linewidth]{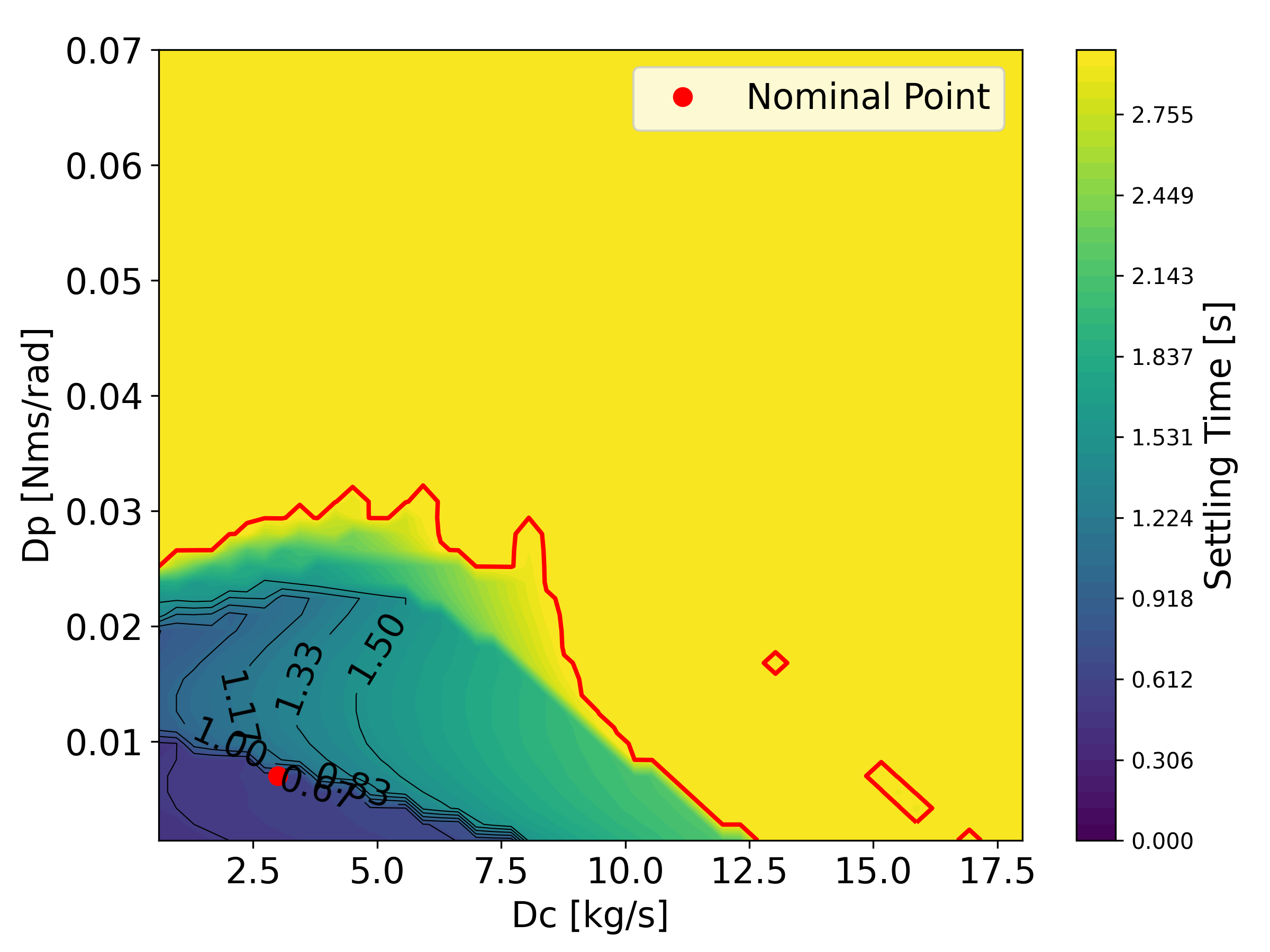}
\caption{Angular 5\% settling-time contours of the GC-DR-RNN across viscous damping parameters \(D_c\) (cart) and \(D_p\) (pole pivot), using nominal context values.}
\label{fig:DD_1x}
\end{figure}

\begin{figure}[!t]
\centering
\includegraphics[width=\linewidth]{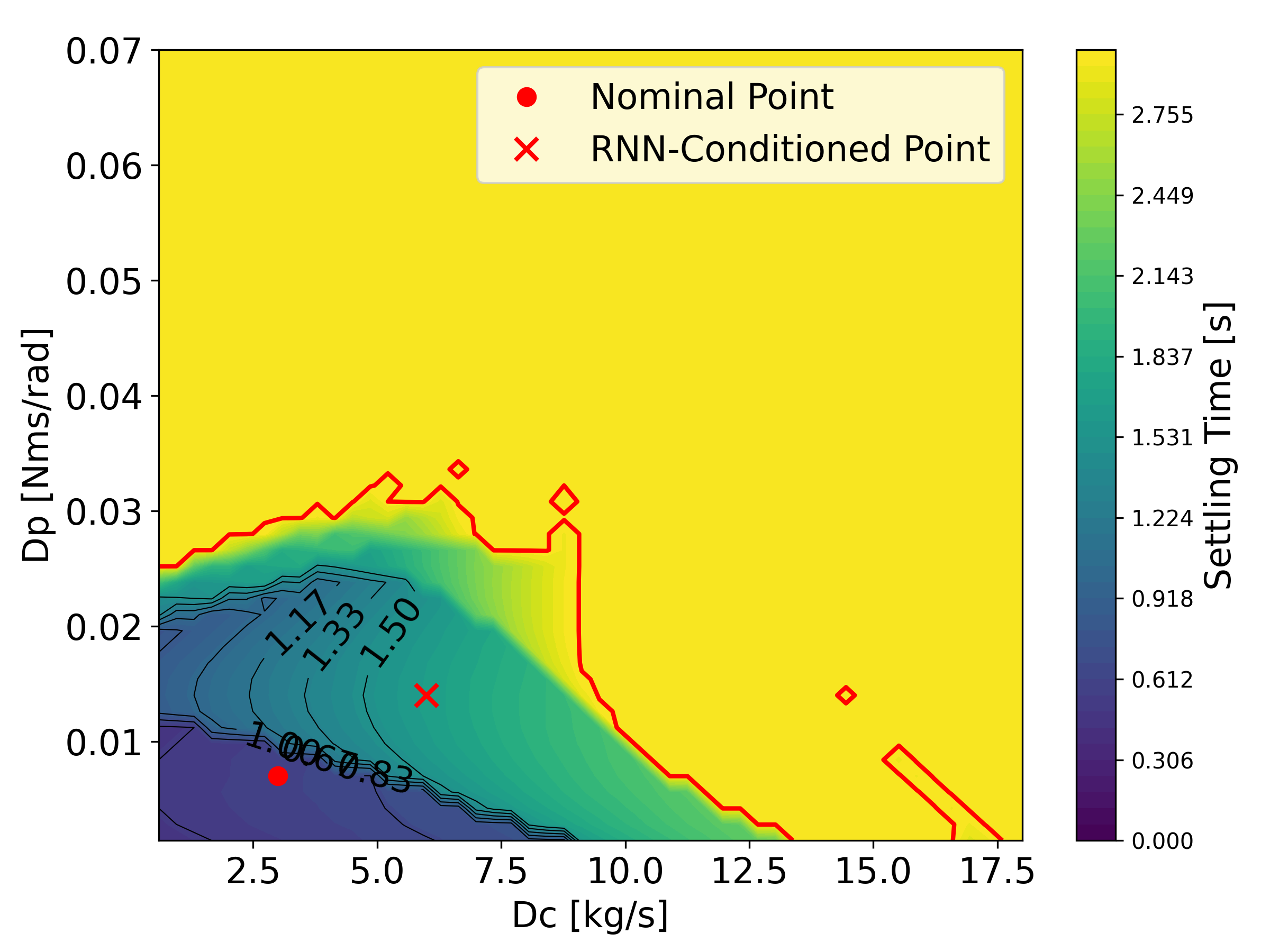}
\caption{Angular 5\% settling-time contours when conditioning the GC-DR-RNN on twice the nominal damping values. The stable region shifts upward but remains centered on moderate damping levels.}
\label{fig:DD_2x}
\end{figure}

\begin{figure}[!t]
\centering
\includegraphics[width=\linewidth]{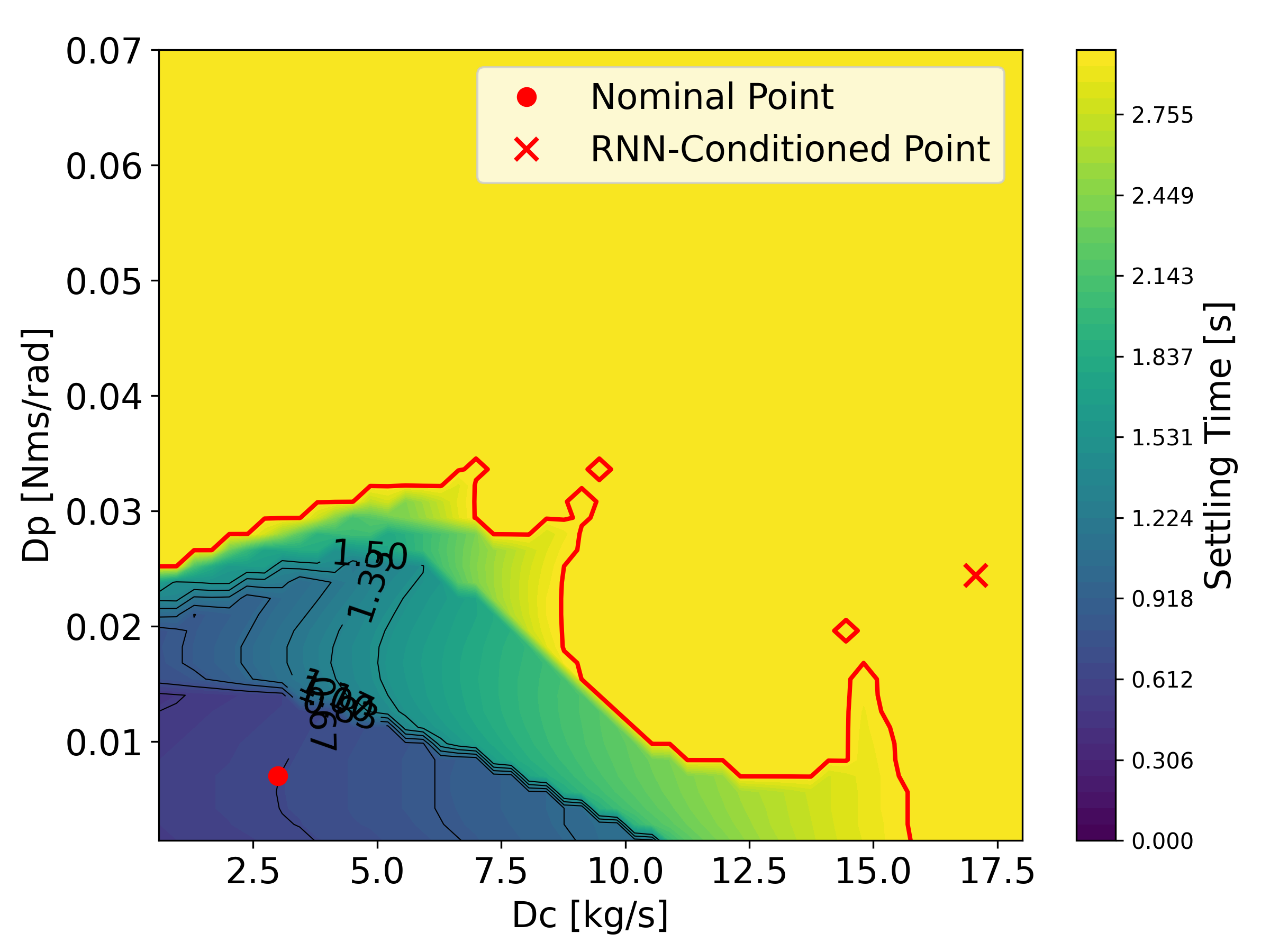}
\caption{Angular 5\% settling-time contours when the GC-DR-RNN is trained with extreme (out-of-domain) damping values. Despite this mismatch, the model achieves a wide stability region.}
\label{fig:DD_far}
\end{figure}

The simulated performances are calculated for each pole and cart mass pair. The contour plots for the proportional controller, the DR-RNN, and the GC-DR-RNN are shown in Fig. \ref{fig:a5Mm_P}, Fig. \ref{fig:a5Mm_DR}, and Fig. \ref{fig:a5Mm_GCDR}.
These plots indicate that domain-randomized RNNs exhibit wider robustness over plant parameters. This observation supports the effectiveness of domain-randomized training, widely used in recent robotics, for mitigating the sim-to-real gap. However, a closer look at robustness around the nominal parameters reveals a drawback of straightforward domain randomization: the stable region near the nominal point becomes narrower.
To investigate the cause of this effect, we visualized the final angular velocity loss and the final mini-batch samples, shown in Fig. \ref{fig:wloss_DR_Mm} and Fig. \ref{fig:wloss_GCDR_Mm}. These visualizations reveal interesting properties of domain randomization, such that the randomization range and the generalization (stabilization) ranges broadly differ. Although the maximum randomized value of $M$ is below 0.8 kg, the minimum-loss point in the DR model lies around 2.0 kg. This visualization implies that the task is ``easier''at large masses, biasing the model toward such easy samples and degrading performance near the nominal plant parameters. The proposed GC model suppresses this tendency and maintains robustness near the nominal parameters, lying between the proportional and DR controllers.
To determine the RNNs' context plant parameters, we varied the friction coefficients, which are difficult to identify accurately. The results are shown in Fig. \ref{fig:DD_1x}, Fig. \ref{fig:DD_2x}, and Fig. \ref{fig:DD_far}. The result is intuitive because the performance landscape is deforming towards the contextualized plant values. However, it is also counterintuitive in that the RNN also generalizes on the out-of-domain plant context values, and the in-domain robustness in the friction plane is best at these out-of-domain plant context values.  Thus, we selected out-of-domain context values of $D_c=17.06351\,\mathrm{kg/s}$, $D_p=0.024376\,\mathrm{Nms/rad}$ for the experiments.


\section{Experimental Results}
\begin{figure}[!t]
\centering
\includegraphics[width=\linewidth]{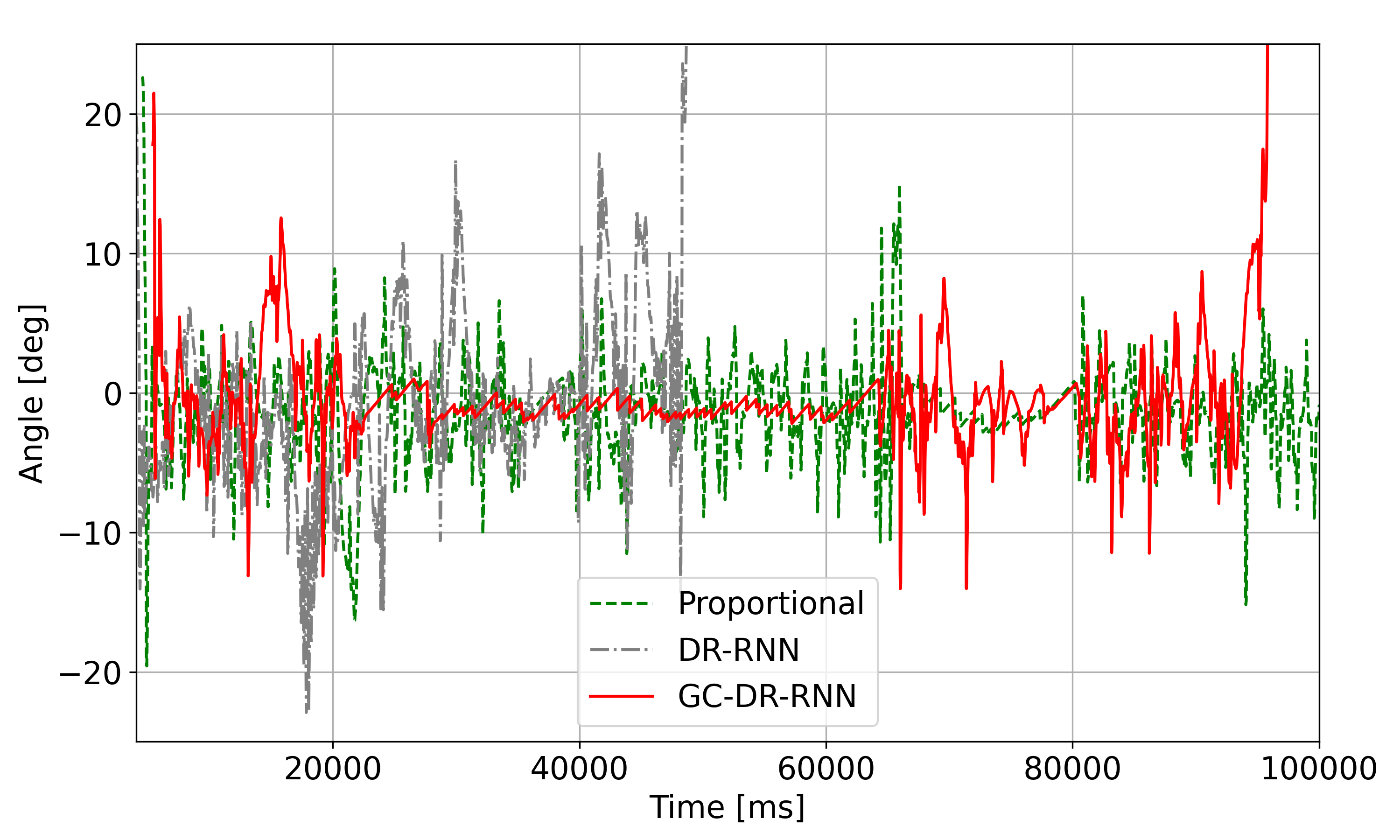}
\caption{Real-world roll-angle responses under three controllers: proportional (green dashed), DR-RNN (gray dash-dot), and GC-DR-RNN (red solid).}
\label{fig:exp_angle_all}
\end{figure}

\begin{figure}[!t]
\centering
\includegraphics[width=\linewidth]{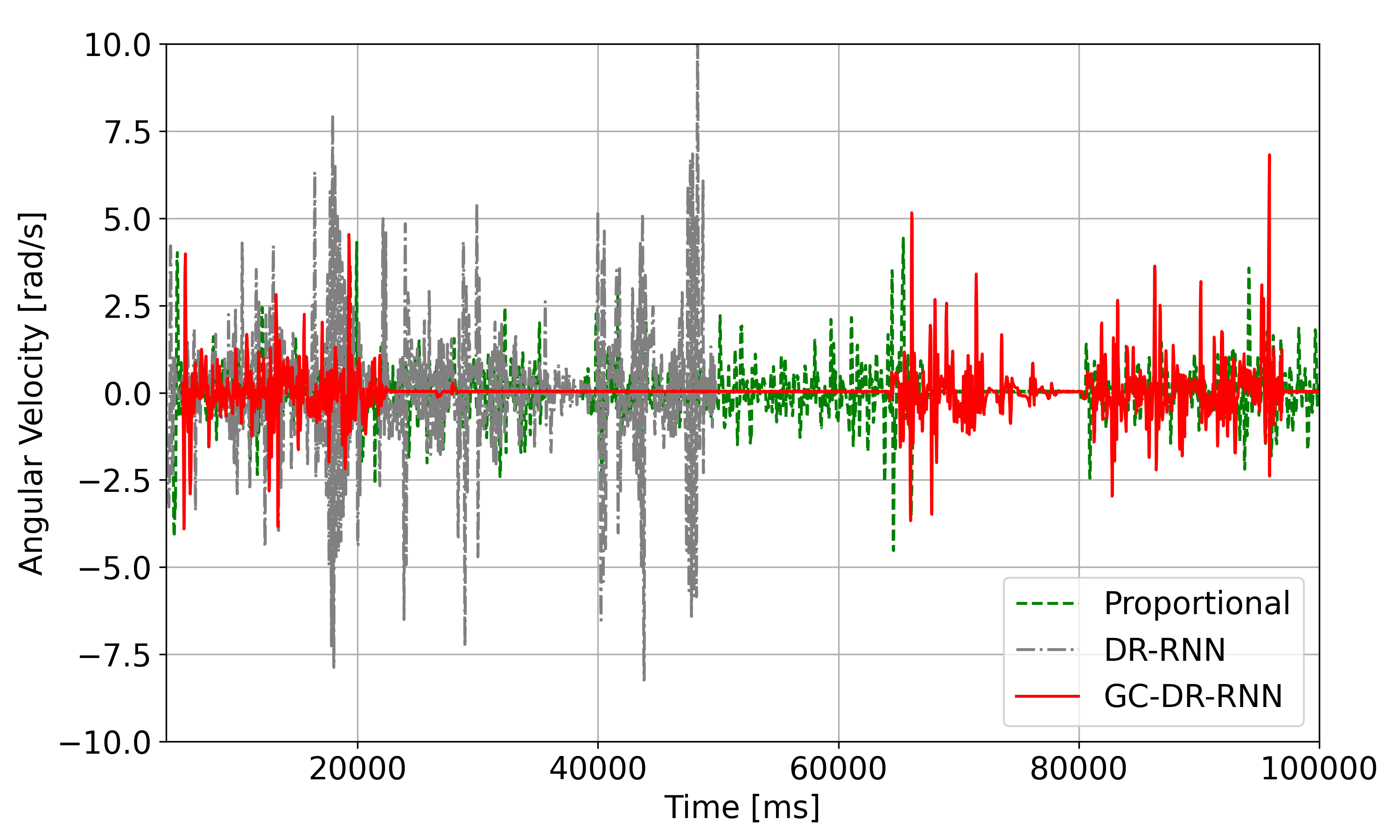}
\caption{Measured angular velocity traces: the DR-RNN (gray) oscillates beyond ±5 rad/s, whereas the GC-DR-RNN and proportional controller almost remain within ±2.5 rad/s and exhibit smooth corrections.}
\label{fig:exp_angvel_all}
\end{figure}

\begin{figure}[!t]
\centering
\includegraphics[width=\linewidth]{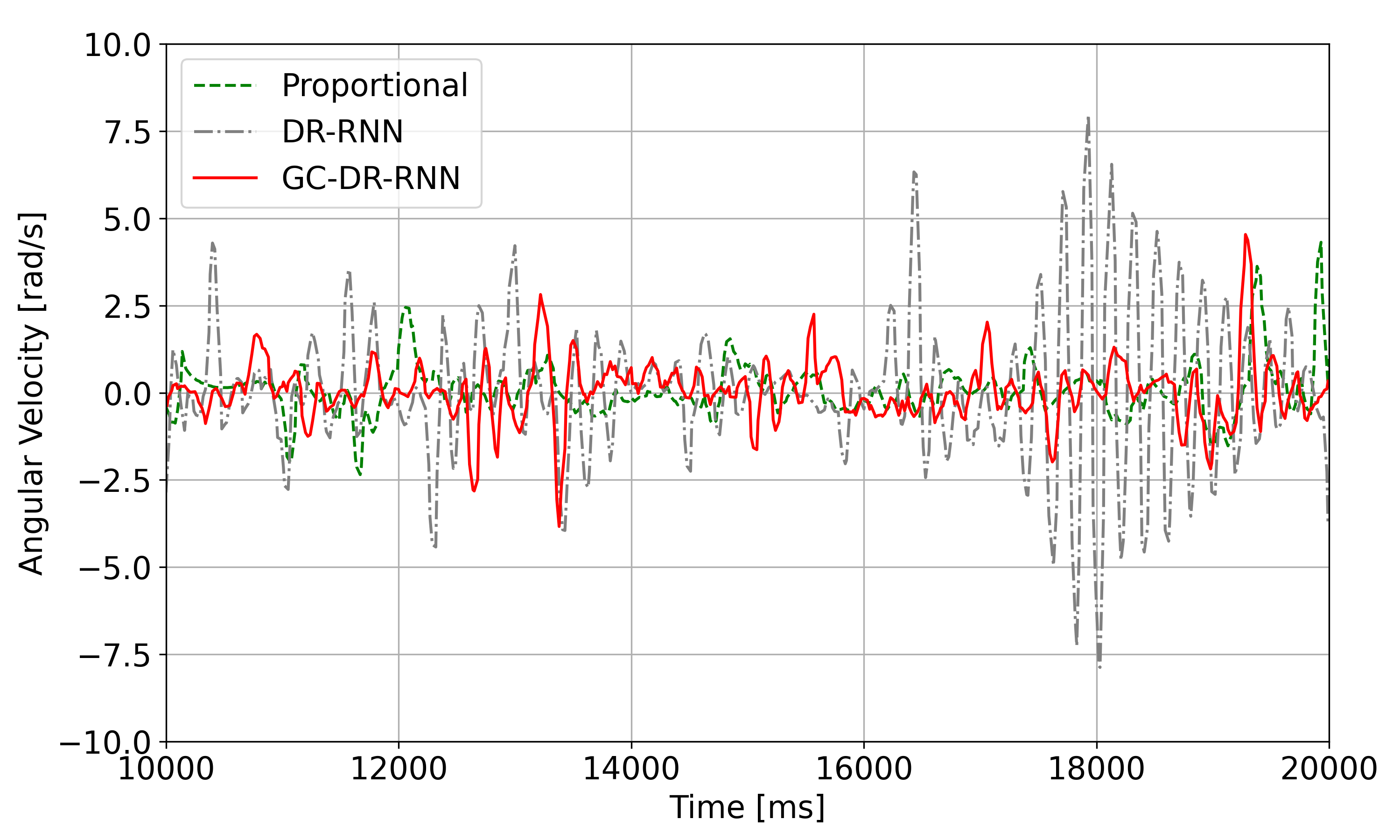}
\caption{Zoom into 10--20 s of angular velocity data. The DR-RNN retains persistent $\simeq 5$ Hz oscillations, absent in both the GC-DR-RNN and the proportional controller.}
\label{fig:exp_angvel_zoom}
\end{figure}

\begin{figure}[!t]
\centering
\includegraphics[width=\linewidth]{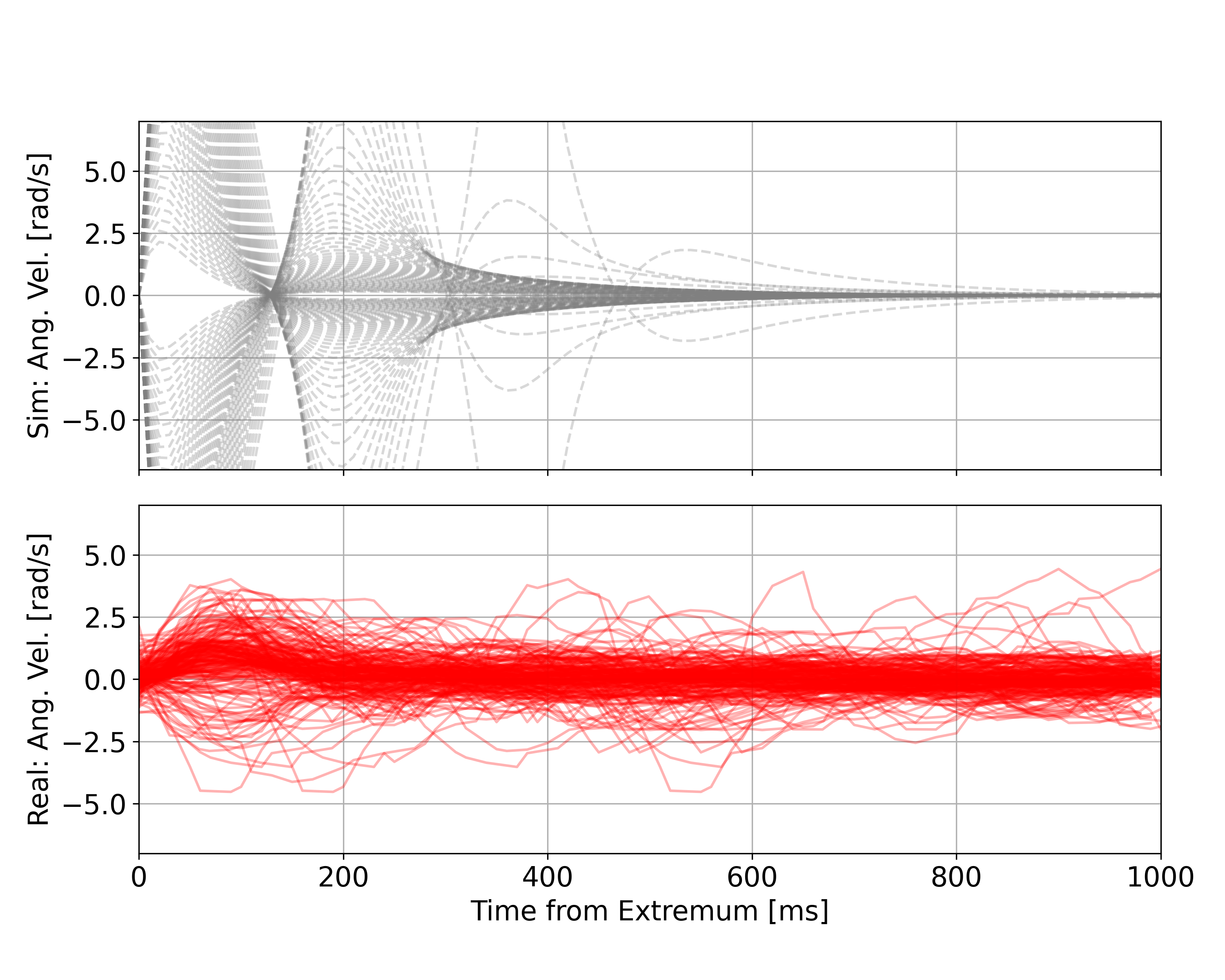}
\caption{Simulated vs. real angular velocity recovery for the proportional controller. Peak times and decay rates align closely, demonstrating high transfer fidelity.}
\label{fig:sim2real_P}
\end{figure}

\begin{figure}[!t]
\centering
\includegraphics[width=\linewidth]
{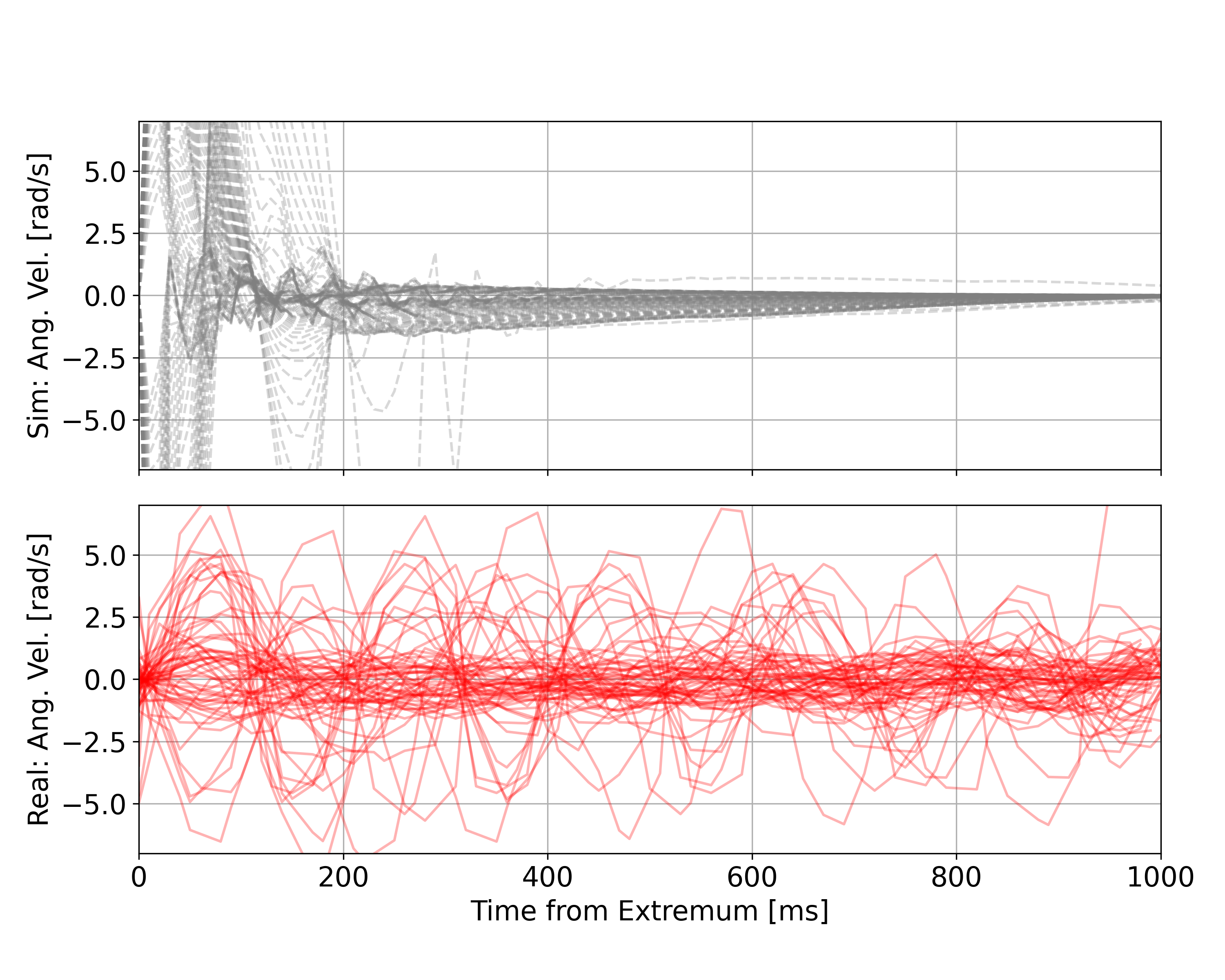}
\caption{Simulated vs. real angular velocity recovery for the DR-RNN: while simulation predicts fast convergence, the real system exhibits clear oscillations, highlighting the sim-to-real mismatch without gain regularization.}
\label{fig:sim2real_DR}
\end{figure}

\begin{figure}[!t]
\centering
\includegraphics[width=\linewidth]{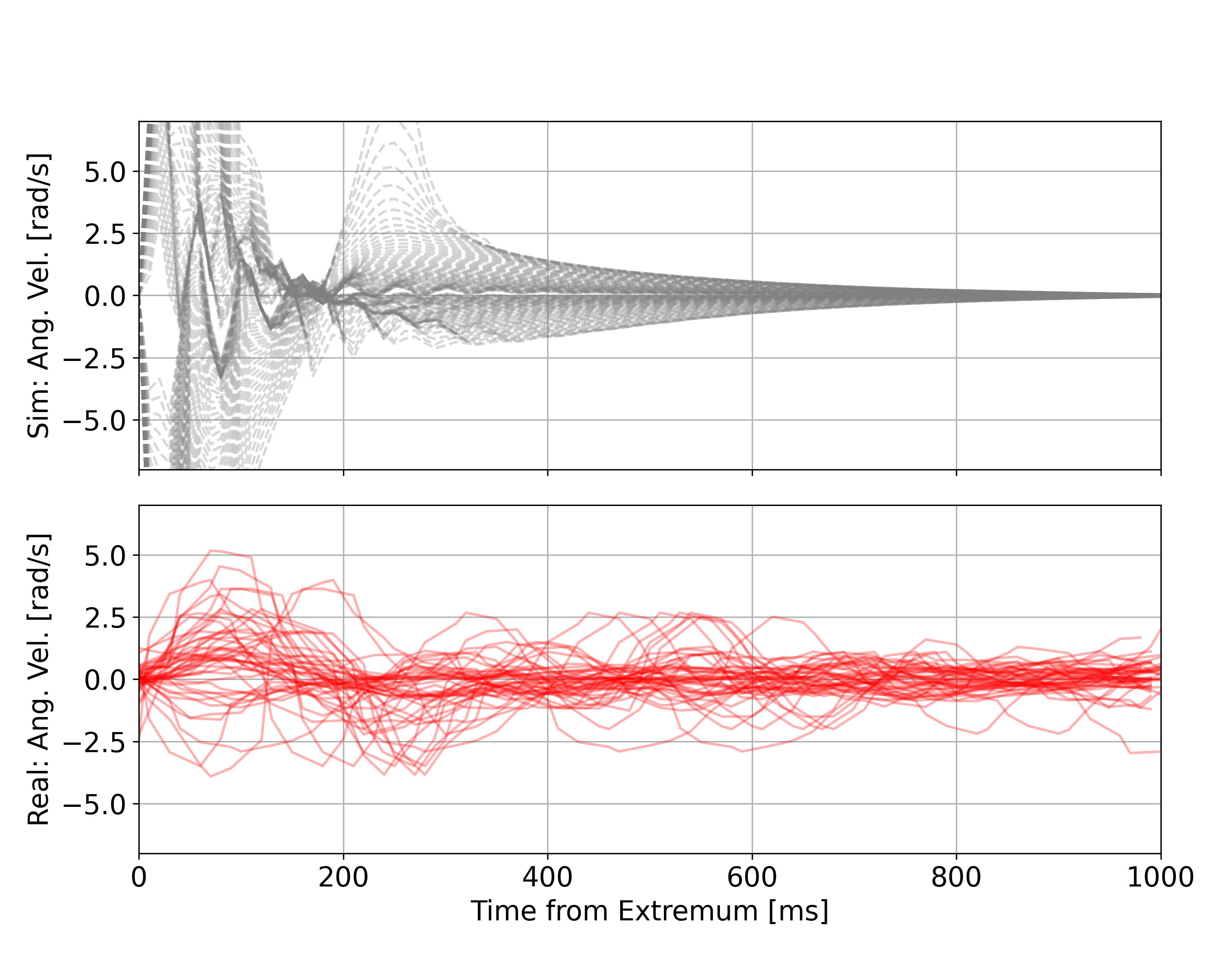}
\caption{Simulated vs. real angular velocity recovery for the GC-DR-RNN: both domains show nearly identical decay envelopes and peak timings, confirming effective sim-to-real transfer with physics-guided regularization.}
\label{fig:sim2real_GCDR}
\end{figure}

The angle and angular velocity records from the balancing experiment are shown in Fig. \ref{fig:exp_angle_all} and Fig. \ref{fig:exp_angvel_all}.\footnote{The first few seconds contain the machine holding time, after that, the machine is self-controlled before it falls (recording stops). Because the machine has relatively high friction in its moving parts, it sometimes becomes completely self-standing without control. We waited for the machine to fall again naturally in such a situation.}
All models exhibit basic balancing capability on the real hardware. However, in the close-up shown in Fig.~\ref{fig:exp_angvel_zoom}, anomalous oscillations appear in the DR-RNN. This phenomenon was also visible to human eyes. This observation is consistent with the proportional gain search in pre-experiment Table \ref{tab:proportional_gains_preexperiment} and the relatively high equivalent angular velocity gain in DR-RNN, which is nearly outside the range of the working proportional gains.
We used a peak-detection algorithm on the angular response of the simulator and real data, and watched the transient recovery responses of the states. We show the aligned angular velocity responses in Fig. \ref{fig:sim2real_P}, Fig. \ref{fig:sim2real_DR}, and Fig. \ref{fig:sim2real_GCDR}.
These figures reveal the sim-to-real gap. Both the proportional controller and the GC-DR-RNN have similar exponentially decaying envelops, and the peak times are almost the same between the simulator and the real machine. However, the DR-RNN shows a clear discrepancy of peak times between the simulator and the real machine. The simulator result shows faster convergence than the other models, but the real response shows oscillation.


\section{Conclusion}
We adopted a simple consumer robot and plant model to expose the sim-to-real gap. We proposed a training constraint that reflects real-machine characteristics through the range of working proportional gains. The visualization in the plant parameter space showed that the domain randomization is effective in the wide robustness area, which may include multiple proportional controller performances in a single RNN. However, domain randomization alone enlarges sim-to-real mismatches around the nominal plant parameters by overemphasizing ``easy'' samples. Our proposed constraint suppresses the side-effect of domain randomization and shows better stability than the only domain-randomized controller on the real machine.

\section{Limitations}

\subsection{Mismatch in conditioned physical parameters}
We append simulated plant parameters (mass, length, damping, etc.) to the controller input, using identical values during training and deployment. However, our experiments revealed that the ``best'' performance did not always occur at the hardware parameters. To use the measured values directly as the RNN condition requires further research.

\subsection{Performance parity with proportional control}
Quantitatively, RNNs' settling times matched those of a well-tuned proportional controller, rather than exceeding them. Although some test runs yielded models that appeared to outperform proportional control in real-world trials, reliably achieving measurable advantages will require further refinement of loss functions or training strategies.

\subsection{Remaining seed sensitivity}
While our GC-DR-RNN shows far less training fragility than a pure RL+DR controller, it still inherits the seed sensitivity of deep networks. In practice, we found that only a few reruns, which filter out classical-control-style instabilities, are sufficient to obtain a stable model.

\section{Acknowledgments}
\noindent I thank my colleague, Mitsuru Ambai, for insightful discussions on this paper. The manuscript uses ``we'' for stylistic consistency. This research was partially supported by OpenAI's ChatGPT, which was employed to reduce routine workload---such as generating source code, composing manuscripts, and improving grammar. The core ideas and research questions were conceived by the author. All AI-generated content was thoroughly proofread and debugged by the author. No AI-generated outputs were used to fabricate experimental results or to mislead readers. The author takes full responsibility for the scientific content presented herein.

\bibliography{main}
\bibliographystyle{abbrv}

\end{document}